# Towards energy-efficient Deep Learning: An overview of energy-efficient approaches along the Deep Learning Lifecycle

**Vanessa Mehlin**
University of Applied Science Ansbach
vanessamehlin@gmail.com

**Sigurd Schacht**
University of Applied Science Ansbach
sigurd.schacht@hs-ansbach.de

**Carsten Lanquillon**
University of Applied Science Heilbronn
carsten.lanquillon@hs-heilbronn.de

## ABSTRACT

Deep Learning has enabled many advances in machine learning applications in the last few years. However, since current Deep Learning algorithms require much energy for computations, there are growing concerns about the associated environmental costs. Energy-efficient Deep Learning has received much attention from researchers and has already made much progress in the last couple of years. This paper aims to gather information about these advances from the literature and show how and at which points along the lifecycle of Deep Learning (IT-Infrastructure, Data, Modeling, Training, Deployment, Evaluation) it is possible to reduce energy consumption.

## 1. Introduction

The last decade has seen tremendous developments in the fields of Artificial Intelligence (AI) and Machine Learning (ML), mainly due to the availability of massive datasets and the advancement of Deep Learning (DL). DL consists of neural networks with multiple hidden layers known as deep neural networks (DNNs) [1]. DNNs are used in various products and services, such as speech assistance, autonomous driving, or facial recognition. They provide superior performance in these tasks compared to traditional models, and there is a trend for developing even larger and more powerful DNNs [2]. However, these models generally require large amounts of data and high computing power, which is associated with high energy consumption incurring high financial and environmental costs [3]. Furthermore, for emerging applications such as autonomous driving and the Internet of Things, models must be run on low-power devices [4]. Therefore, energy-efficient DL is crucial from an economic, environmental, and application perspective.

The efficiency of DL can be divided into training efficiency and inference efficiency [5]. Model training comes at a high environmental cost, as energy is required to run it on hardware for weeks or months at a time [3]. However, training a DL model is only the beginning of the lifecycle. Once the model is trained, it will be implemented and used. This process, called inference, also consumes enormous energy. Inference does not last weeks or months, but unlike training, it is not a one-time event. It takes place continuously and can therefore exceed the energy consumption of the training after a certain number of inference events [6].

Various researchers have presented approaches, methods, and techniques such as mixed-precision training, pruning or knowledge distillation that can accelerate training and inference time and reduce the energy consumption of the models. This paper aims to provide an overview of these existing approaches and categorize them into the phases of the DL lifecycle. The lifecycle considered in this paper encompasses IT-Infrastructure, Data, Modeling, Training, Deployment (Inference) and Evaluation. The subdivision of the lifecycle is based on the Data Science Process Model (DASC-PM) [7] as well as on the CRISP-DM [8], which is the Cross Industry Standard Data Mining Process Model. To the authors' knowledge, there has been no previous work aimed at providing a global overview of existing energy-reducing approaches along the lifecycle of DL. There have been overviews of energy-efficient DL ([5], [9]–[12]); however, these overviews have neglected the evaluation part of the DL lifecycle. Thus, this paper is intended not only as a holistic overview of energy-efficient approaches along the lifecycle of DL, it also provides new insights beyond the information provided by these previous overviews.

The paper is structured as follows. Section II presents the overviews that served as the foundation for the present work. This is followed by Section III, in which the methodology and research questions are outlined. Section IV provides an overview of the various approaches to reducing energy consumption in the different phases of the DL lifecycle. Finally, a conclusion and a discussion on future work are given.

## 2. Related Work

A great deal about energy-efficient DL has been written, including approaches to reduce energy consumption [13]–[19], studies about how much energy DL is consuming ([3], [20], [21]), publications that point out the need for efficiency improvements due to carbon emissions or the use of edge devices ([22], [23]) as well as the following overviews of energy-efficient DL:

Menghani [5] gave an overview of efficient modeling techniques, infrastructure and hardware. A collection of algorithms, techniques, and tools related to efficient DL in the five areas of (1) compression techniques, (2) learning techniques, (3) automation, (4) efficient architectures, and (5) infrastructure are presented. Furthermore, an experimental guide with code to show how these tools and techniques work with each other is provided. This guide helps practitioners to optimize model training and deployment

Xu et al. [9] conducted a systematic review of energy-efficient DL technologies within the four categories: (1) compact networks, (2) energy-efficient training strategies, (3) energy-efficient inference approaches, and (4) efficient data usage. For each category, they provided a taxonomy. Furthermore, they discussed the progress made and the unresolved challenges in each category.

Cai et al. [10] and Liu et al. [11] focused on memory and compute-limited devices (edge/mobile devices). They surveyed DL approaches including lightweight network design, network compression, hardware-aware neural architecture search, adaptive models, efficient on-device learning, and efficient system design.

Lee et al. [12] provided an overview of resource-efficient DL techniques in terms of (1) modeling, (2) arithmetic, and (3) implementation techniques. The focus of this overview is on resource-efficient techniques for Convolutional Neural Network architectures, since it is one of the most commonly used DL architectures.

## 3. Methodology

To provide a holistic overview of energy-efficient approaches along the lifecycle of DL an initial literature review on existing overviews and literature reviews were conducted. As a result, four recent overviews of energy-efficient DL ([5], [9]–[12]) were found (see Related Work). Two of those ([10], [11]) focus on mobile/edge devices and one of those [12] focuses on resource efficiency. DL on edge and mobile devices are closely related topics, since those devices have limited memory and compute resources, depending on energy efficiency [10]. Resource-efficient DL is also strongly related, as some resource-efficient techniques, such as mixed-precision training are associated with energy efficiency. However, in the further course of this work, these edge/mobile- and resource-focused overviews are not considered to the same extent as the other two surveys, due to (1) a focus on the energy efficiency of DL in general and (2) due to the limited scope of the study. Since the remaining two overviews ([5], [9]) together already comprehensively cover the topic, it was decided to synthesize the collected information from these two overviews as a basis rather than conduct another individual literature review on energy-efficient DL.

The next step was to verify whether the techniques and approaches gathered in the overviews cover the entire DL lifecycle (table 1). The key takeaway was that the reviewed overviews neglect the evaluation part of the DL lifecycle. The topic is addressed, but no such overview is provided as in the other categories. Since the evaluation part is not equally represented, it was decided to guide this paper with the two research questions and methods presented in Table 2.

Tab. 1: Mapping of the content of the overviews along the DL lifecycle
* = it is addressed, but no such overview is provided as in the other categories

| Overview | IT-Infrastr. | Data | Modeling | Training | Deployment | Evaluation |
|---|---|---|---|---|---|---|
| Menghani [5] | ✓ | ✓ | ✓ | ✓ | ✓ | * |
| Xu et al. [9] |  | ✓ | ✓ | ✓ | ✓ | * |

Tab. 2: Research questions and methods

| No. | Research question | Method |
|---|---|---|
| RQ1 | Which techniques and approaches can be applied in the phases of the DL lifecycle to reduce the energy consumption of DL models? | Synthesizing information from related overviews and mapping them along the DL lifecycle. |
| RQ2 | How can models be evaluated in terms of their energy consumption and carbon emissions? | (1) Synthesizing information from the related overviews (2) Literature review |

To answer RQ1 the techniques and approaches of the two reviewed overviews were analyzed and then classified into a suitable phase of the DL lifecycle. As mentioned by Xu et al. [9] it can be challenging to provide an overview of energy-efficient DL due to the lack of a unified standard measurement for energy-efficient DL. Therefore, it is sometimes difficult to tell whether an approach is energy-efficient or not. One controversial example is Neural Architecture Search Algorithms (explained in section 4). Running such algorithms usually needs large computational resources [5]. But the resulting efficient models can significantly reduce the computational burden in downstream research and production, leading to reduced energy consumption [24]. Therefore, it is debatable whether a technology is defined as energy-efficient or not. However, if an approach has the potential to reduce the energy costs of DL models, it is included in the overview. Furthermore, a clear classification of some approaches in a certain phase of the DL lifecycle is not given. In some cases, an approach could be assigned to two or three phases, depending on the interpretation. For example: Pre-trained models could be mapped in the "Data", "Modeling" or "Training" phase. Thus, the classification of some approaches depends on the author's interpretation, which is based on the categorization of the two studies reviewed. Not all of the techniques listed in the overviews will be presented, because some of them are just mentioned within the overview and not described as for example "K-Means Clustering" in Menghani [5]. Additionally, some of them are not relevant in the same extent as the others, for example Hyper-Parameter Optimization. It is mentioned in Menghani [5], but in the remainder of this paper, the focus is on Neural Architecture Search, as this is the most recent advance in the field. To answer RQ2 a literature review was conducted. The search was performed during May 2022 and the meta search engines Web of Science, Scopus as well as Google Scholar, which cover all major publishers and journals such as Springerlink, Elsevier, IEEE Xplore and Research Gate, were used. Searches utilizing the following search terms were performed across all databases:

("deep learning" OR "DNN" OR "deep neural network" OR "machine learning") AND ("measure*" OR "metric" OR "evaluation" OR "report*") AND ("environment" OR "energy" OR "emission" OR "footprint")

These search strings were modified for the different databases, using specific search functions for each database. Using this search strategy, 78 journal articles were identified for further analysis. After reading abstract and full text 14 relevant articles remained to answer RQ2.

## 4. Findings

RQ1: *Which techniques and approaches can be applied in the phases of the DL lifecycle to reduce the energy consumption of DL models?*

Table 3 summarizes the different energy-efficient techniques and approaches, extracted from the overviews, at each stage of the DL lifecycle. In the following, each phase will be presented in more detail (tables 4 to 8). Since this paper aims to provide a holistic overview of which energy-efficient methods exist and in which phase of the DL lifecycle they can be used for more energy efficiency, the different approaches are not described in detail. A detailed description can be found in the additional references (Add. Ref.). These references are either collected from the overviews or included by the author.

Tab. 3: Energy-efficient techniques and approaches along the DL lifecycle

| IT-Infrastr. | Data | Modeling | Training | Deployment |
|---|---|---|---|---|
| Software<br>•Tensorflow<br>• PyTorch<br>• Hardware-optimized libraries | Active Learning | Design:<br>• Compact Convolution<br>• Efficient Attention<br>• Lightweight Softmax<br>• Compact Embedding | Initialization | Pruning |
| Hardware<br>• GPU<br>• TPU | Data Augmentation | Assembling:<br>• Memory Sharing<br>• Static Weight Sharing<br>• Dynamic Weight Sharing | Normalization | Low-Rank Factorization |
| | | | | Quantization |
| Neuromorphic Computing | Pre-trained models | Hyper-Parameter Optimization/Neural Architecture Search | Progressive Training | Knowledge Distillation |
| | | | Mixed-Precision Training | Deployment Sharing |

## 4.1 IT-Infrastructure

The basis for running a DL model efficiently is a robust software and hardware infrastructure [5]. This section provides an overview of software and hardware components that are critical to model efficiency.

Tab. 4: Efficient IT-Infrastructure

| Technique/ Approach | Description | Add. Ref. |
|---|---|---|
| Software | • Tensorflow is a machine learning framework which has some of the most extensive software support for model efficiency (e.g. TF Lite, TF Model Optimization toolkit) [5].<br>• PyTorch is a machine learning framework, which includes for example PyTorch Mobile (light-weight interpreter that enables running PyTorch models on mobiles) and a model tuning guide that lists various options available to practitioners such as mixed-precision training or enabling device-specific optimizations [5].<br>• Hardware-optimized libraries: Efficiency can be further increased by optimizing for the hardware on which the neural networks run [5]. | [25]<br><br>[26] |
| Hardware | • GPU (Graphics Processing Units) were originally used for computer graphics, until a study in 2009 [27] demonstrated that it can be used to accelerate DL models [5].<br>• TPUs (Tensor Processing Units) are proprietary application-specific integrated circuits (ASICs) that Google developed to accelerate DL applications using Tensorflow [5]. | [28] |
| Neuromorphic Computing | Neuromorphic Computing refers to a variety of computers, devices, and models that are inspired by the brain and contrast the widespread von Neumann computer architecture [29]. Schuman et al. 2017 [29] discovered that one of the main motivations for neuromorphic computing is speed of computation and their potential for extremely low power operation. Especially the developers of early systems highlighted that it is possible to perform much faster neural network computations with customized chips than with traditional von Neumann architectures, in part by exploiting their natural parallelism, but also by building customized hardware for neural computations. This early focus on speed was a forerunner of the future of using neuromorphic systems as accelerators for machine learning or neural network style tasks [29]. | [29]<br>[30] |

## 4.2 Data

It is common to increase training data to achieve better model performance. The downside of this approach is the significant increase in training costs [9]. This is one of the reasons why data efficiency has received significant attention over the years [22]. In this phase three techniques are presented that can be used to gain data efficiency, accelerate training time, and achieve competitive results with fewer data resources.

Tab. 5: Efficient Data Usage

| Technique/ Approach | Description | Add. Ref. |
|---|---|---|
| Data Augmentation | Data augmentation is a solution to address the scarcity of labeled data. The basic idea is to synthetically inflate the existing dataset through augmentation methods, so that the new label of the augmentation example does not change or can be derived in a cost-effective way. An example of this is the classic dog or cat image classification task. If the image of a dog is shifted horizontally/vertically by a small number of pixels or it is rotated by a small angle, the image does not change significantly, so the transformed image should still be classified as "dog". As a result, the classifier is forced to learn a | [31]<br>[32] |

| Active Learning | Active learning aims to achieve good results with as few samples as possible. It was originally proposed to reduce annotation costs. Today, pool-based active learning is widely used to reduce training costs by selecting the most useful samples for training a network. The idea behind active learning is the following: Annotated training data do not contribute equally to final performance. Thus, if only the most useful sample is selected for training models, the waste of training on irrelevant samples can be largely reduced. [9] | [33] [34] [35] [36] |
|---|---|---|
| | robust representation of the image that is more generalizable across these transformations. [5] | |
| Pre-trained models | Pre-trained models as initialization can be an effective approach to reduce data requirements in downstream tasks [5]. This approach is further discussed in the training phase. | [37] |

## 4.3 Modeling

Continuous improvements in model architectures lead to significant reductions in computational effort required to achieve a given level of accuracy. For example, the Transformer architecture developed in 2017 required 10 to 100 times less computational effort while achieving better results than the state-of-the-art models at the time [13]. This chapter concentrates on such efficient neural networks. Furthermore, assembling and automation as presented in Xu et al. [9] will be addressed.

Tab. 6: Efficient Modeling

| Technique/ Approach | Description | Add. Ref. |
|---|---|---|
| Efficient Architecture Design | | |
| Compact Convolution (Vision) | In the following compact convolution methods that improve resource-efficiency are listed: <br> • Depthwise separable convolution <br> • Downsampling <br> • Flattened convolution <br> • Group convolution <br> • Linear bottleneck layer <br> • Octave convolution <br> • Shrinked convolution <br> • Squeezing channel/fire convolution | [38] [39] [40] [41] [42] [43] [44] [45] |
| Efficient Attention | The attention mechanism directly aligns all tokens from sequence-to-sequence models together, which can address long-distance dependencies to some extent. But due to the fact that any two tokens have an attention score, the required computations grow quadratically with the input length [9]. To address this problem, several studies proposed efficient attention variants. Xu et al. [9] classified them into the following categories: <br> • Sparse attention: Reduces the span of attention <br> • Attention approximation: Different attention estimation formats | [16] [46] [47] [48] [49] |
| Lightweight Softmax (NLP) | Softmax layer introduces embeddings for all tokens, which leads to many computations for a large vocabulary [9]. Therefore, several efficient lightweight softmax variants have been proposed [9]. Xu et al. [9] distinguished these variants as follows: <br> • Reduction of parameters: The proposal is to create a sequence at character level instead of word level. The number of characters is much smaller than that of words, which helps to reduce the computations for softmax significantly. <br> • Reduction of computations: Xu et al. [9] classified softmax variants with fewer computations into five categories: <br>     o Hierarchical softmax <br>     o Softmax with dynamic embeddings <br>     o Sampling-based softmax | [50] [51] [52] [53] [54] [55] [56] |
| | o Hashing-based softmax<br>o Normalization-based softmax | |
| Compact Embedding (NLP) | The first step for NLP tasks is building token embeddings. Reducing the parameters of these embeddings to make them compacter is an important topic. There are several approaches to compress neural networks such as pruning, knowledge distillation, low-rank approximation, and quantization. Xu et al. [9] divided approaches for compact embeddings into four categories:<br>• Reuse-based approaches<br>• Knowledge-distillation-based approaches<br>• Low-rank-based approaches<br>• Fine-grained vocabularies | [57]<br>[58]<br>[59]<br>[60]<br>[61]<br>[62] |
| Assembling | Component assembling solutions aim for efficient architecture design. The key idea of efficient component assembling lies in sharing [9]. | |
| Memory Sharing | Memory sharing is a technique for storing large models on devices with limited memories (IoT). The idea is to share the same storage among intermediate forward vectors [63] or backward vectors [64] to reduce memory requirements [9]. | [65]<br>[66] |
| Static Weight Sharing | Static weight sharing aims to explore how weights can be reused for a neural network. The weights are fixed during inference and shared among all samples. To save memory, many models choose to reuse parameters across different layers or different tasks (dealing with problems involving multiple tasks, multiple domains, or multilingual problems) [9]. | [67]<br>[68]<br>[69]<br>[70] |
| Dynamic Weight Sharing | Static parameter sharing usually fails when dealing with tasks that are not closely related. To decide which layers/components should be shared by different input samples dynamic solutions can be used. Dynamic networks are neural networks with dynamic computational graphs in which the computational topology or parameters are spontaneously determined to reduce computation costs and improve the adaptiveness of networks [9]. Xu et al. [9] provided an overview of general dynamic architectures. Within the cascading-style network architectures, for example, first smaller networks and then larger ones are executed. If a smaller network can already process the input sample, the model stops the execution process and does not execute any further models. Skipping-style networks speed up inference by either skipping certain layers or omitting unimportant input spans in the entire input sequence. Other general dynamic architectures are early-exit-style networks and mixture-of-experts-style networks [9]. | [71]<br>[72]<br>[73]<br>[74]<br>[75]<br>[76]<br>[77]<br>[78] |
| Automation | Automated approaches can automatically search for ways to train more efficient models. The downside is that these methods may require large computational resources [5]. However, while the initial effort can be computationally intensive, the resulting efficient models can significantly reduce the computational effort required in downstream production, resulting in lower overall energy consumption [24]. | |
| Neural Architecture Search (NAS) | NAS is a technique to search automatically for a global optimal efficient DNN model. But it is very time-consuming and computationally expensive [79]. To solve this problem, some new NAS algorithms are proposed [5]:<br>• Evolution based search reduces the search costs. It is a two-stage search approach, where several well-functioning parent architectures are selected in the first stage. In the second stage mutations are applied on these architectures to select the best one. For this stage pre-trained parent networks which does not require too many computations to train child networks are used. This method requires validation accuracy as search criterion, which still makes it computationally expensive.<br>• Differentiable search: This method is proposed to completely eliminate the dependency on validation accuracy with re-formulating the task in a differentiable manner and allowing efficient search using gradient descent.<br>• Another research direction aims to represent a model in a continuous space where there is a mapping between structures and outcomes. By doing so, the model only learns how to predict the performance of architectures based on their continuous representations where the downstream training is not needed.<br>• Training-free NAS approaches: These approaches directly extract features from randomly-initialized models and use these features as evaluation criterion to select optimal networks. | [80]<br>[81]<br>[82]<br>[83]<br>[84]<br>[85]<br>[86]<br>[17] |

## 4.4 Training

In this section, the focus is on the energy efficiency of training DL models. Many approaches have been proposed to reduce training costs, including initialization, normalization, progressive training, and mixed-precision training.

Tab. 7: Efficient Training

| Technique/ Approach | Description | Add. Ref. |
|---|---|---|
| Initialization | Pre-trained models from other domains (or other tasks) can be used for initialization. It is generally assumed that initialization from existing models can improve the generalization ability with fewer training iterations [9]. Xu et al. [9] categorize these pre-training initialization approaches as follows:<br>• Feature-based initialization: The parameters (usually from the lower or middle layers) are borrowed from other domains/tasks as initialization.<br>• Fine-tuning-based initialization: Here the target data is used to train all parameters (including new parameters and borrowed parameters). It can further optimize the target objectives by fine-tuning all parameters to better fit the training data.<br>• Supervised initialization: This approach is popular in low-resource environments and is extensively studied on domain adaptation/transfer learning. As a common solution, the target model is pre-trained using similar tasks/datasets and then the pre-trained parameters are used as initialization for the target task.<br>• Self-supervised initialization: To reduce the requirements of supervised data, previous studies have dealt with self-supervised pre-training, which uses unlabeled data to construct supervision signals to learn representations. Since self-supervised pre-training does not require human annotated labels, it is easy to obtain sufficient training data. | [87] [88] |
| Normalization | Normalization is based on a special component which can accelerate convergence and is used to normalize hidden outputs in deep neural networks. There are different normalization variants that have almost the same calculation process but are applied to different dimensions or objectives, such as batch normalization, layer normalization, group normalization and weight normalization. Ioffe & Szegedy 2015 [89] were able to perform 14 times fewer training steps with the same accuracy on a state-of-the-art image classification model when they used batch normalization. ([9], [90]) | [90] [91] |
| Progressive Training | The idea of progressive training is to add layers constructively. When compared to full training, progressive training does not require full gradients to all parameters and can therefore reduce the computations required for training. Furthermore, the well-trained lower layers also accelerate the training of the higher layers [9]. | [92] [93] |
| Mixed-Precision Training | A distinction can be made between a simple precision format (FP32), which requires 32 bits of memory, and lower precision format (FP16), which requires 16 bits of memory. The lower precision offers numerous advantages, such as the ability to train and deploy larger neural networks, the reduced load on memory, which speeds up data transfer operations, and finally, the acceleration of computations. However, training with low precision also affects the accuracy of the model: The fewer bits used, the lower the accuracy. Mixed precision training can be used as a solution. This approach can reduce the memory requirements by almost half while maintaining the model accuracy. This is achieved by identifying the steps that require full precision to use FP32 for them, while using FP16 for all other steps. ([12], [94]) | [95] |

## 4.5 Deployment (Inference)

State-of-the-art neural network models have millions of parameters that increase the trained model's size and thus increase inference costs [1]. Therefore, it is important to compress and accelerate these models before deploying them without compromising model accuracy. This section describes common model compression methods for reducing inference costs, including pruning, low-rank factorization, quantization, and knowledge distillation. These techniques are commonly used to compress a model after the training to accelerate inference [1]. Deployment sharing is also presented in this phase.

Tab. 8: Efficient Inference

| Technique/ Approach | Description | Add. Ref. |
|---|---|---|
| Pruning | Pruning can be used to remove redundant elements in neural networks to reduce the size of the model and the computational cost. The main idea is to create a smaller network by removing unimportant weights, filters, channels, or even layers from the original DNN while keeping the accuracy ([9], [96]). Generally, network pruning can be divided into (1) pruning of connections (weights) (unstructured pruning) and (2) pruning of filters or channels (structural pruning) [96].<br>• Very effective for reducing the number of parameters in DNNs [5]<br>• Requires iteratively scoring weights and re-training the network for many iterations [5]<br>• Often leads to non-negligible performance drop when applied to powerful DNNs [5] | [19] |
| Low-Rank Factorization | This techniques uses tensor/matrix decomposition to reduce the complexity of convolutional or fully connected layers in deep neural networks [10]. The aim of low rank factorization is to factorize a weight matrix into two matrices with low dimensions [1].<br>• Reduces memory storage and accelerates DNNs [1]<br>• In comparison to other common compression methods, it can effectively reduce the size of models with a large compression ratio while maintaining good performance [9]<br>• Computationally complicated [9]<br>• Less effective in reducing computational cost and inference time than other common compression methods [9] | [18] |
| Quantization | Quantization reduces computation by reducing the number of bits per weight. It minimizes the bit-width of data storage and flow through the DNN. Computing and storing data with a smaller bit-width enables fast inference with lower energy consumption [96].<br>• Can help significantly reduce model size and inference latency [5]<br>• Easy to implement [9]<br>• Post-training quantization often causes a non-negligible drop in performance, whereas quantization-sensitive training can effectively reduce the performance drop [9] | [97] |
| Knowledge Distillation | The knowledge gained from a large-scale high performing model (teacher model), which generalizes well on unseen data, is transferred to a smaller and lighter network known as the student model [1].<br>• Improves resource efficiency [1]<br>• Same/comparable performance as the larger model [1] | [98] |
| Deployment Sharing | The optimal neural network architecture varies significantly with different hardware resources [14]. Therefore, developing elastic or dynamic models that meet different constraints is crucial for practical applications. During inference, the appropriate subnetwork is selected to satisfy the resource constraints from different devices. By amortization of the one-time training cost, the total cost of the specialized design can be reduced [9]. | [99] [100] |

RQ2: *How can models be evaluated in terms of their energy consumption and carbon emissions?*

Even though more and more research is done in energy-efficient DL, standardized metrics of models' energy efficiency and carbon emissions still do not exist till do not exist [9]. However, many studies encourage researchers to consider these two values when evaluating models ([3], [22], [101], [102]). This would help promote those models that achieve high accuracy with lower energy consumption. It would also raise practitioners' awareness of their carbon emissions, which could encourage them to take active steps to reduce them. To determine the evaluation possibilities of DL models in terms of their energy consumption and carbon emissions, (1) information from the relevant overviews was collected and (2) a literature review was conducted. The found insights and publications can be categorized into "Metrics" and "Tools" as shown in table 9.

Tab. 9: Evaluation Metrics and Tools

| Evaluation metrics and tools | Description |
|---|---|
| **Metric** | |
| Carbon emission | Taking carbon emissions as a metric would be the most logical, since this is what is to be minimized. However, it is difficult to measure the exact amount of carbon released by training or running a model, as this number is highly dependent on the local power infrastructure. Therefore, this metric would not be comparable between researchers at different locations or even at the same location at different times [22]. To measure the carbon emission various online tools (see section "tools") can be used. |
| Run Time | Measuring the running time of training and inference could be a valuable metric if all models had the same hardware and software settings. Since this is not the case and running time is highly dependent on infrastructure settings, models running with different settings are not comparable. Nevertheless, reporting the running time can be important for a general understanding of the impact. ([9], [22]) |
| Energy | The energy consumption can be calculated by multiplying the maximum power consumption of the individual hardware used according to their technical specifications by the number of hardware used for training and then by the training time in hours [20]. |
| Accuracy per Joule | This metric captures the accuracy, latency, and energy tradeoffs between models [12]. Hanafy et al. [103] propose to compute it as follows:<br>*Accuracy/Joule = ModelAccuracy/Energy_per_Request.*<br>It indicates how much energy is required for one unit of accuracy and serves as a normalized metric and a benchmark for comparison between two or more models. |
| Full-Cycle Energy Consumption Metric "Greeness" | This metric proposed from Li et al. [104] focuses on the energy consumption during the training and inference, the accuracy, as well as the model usage intensity. It is calculated as follows:<br>$G(MUI) = Acc\tau/(MUI * IEC + TEC)$<br>• Train Energy Cost (TEC) calculates the total energy consumption of efficient DL throughout the training phase, including base model training, model compression, and retraining the model.<br>• Inference energy cost (IEC) refers to the energy consumed to perform inference for one time.<br>• Model Usage Intensity (MUI) is defined as the average number of inferences in each lifecycle. The importance of TEC and IEC varies depending on the MUI. If an AI system uses the model intensively and the number of inferences in a lifecycle is large, then IEC accounts for a large portion of the energy consumption and conversely.<br>• Accuracy (Acc) refers to the accuracy of the model on a given task. Efficient DL models usually trade accuracy for efficiency and their accuracy degradation can vary significantly. For this reason, the accuracy of the models should also be considered. |
| Model Size/Number of parameters | Model size/Number of parameters are also a crucial factor when it comes to training and inference costs [9]. It can be determined quite easily and it is usually directly correlated with the complexity of the computations [20]. Unlike the previously described measures, it is not dependent on the underlying hardware. Nevertheless, different algorithms use its parameters differently, for example by making the model deeper or wider. As a result, different models with a similar number of parameters are often not comparable [22]. |
| FLOPs | Floating point operations (FLOPs) count the number of operations required to run a model when executing a specific instance. The advantage of this metric is that it is |

|  | nearly independent of hardware and software settings and can therefore be an easy way to make a fair comparison between different models [9]. |
|---|---|
| **Tools** |  |
| eco2AI | eco2AI [105] is an open-source Python library for carbon emission tracking. It allows users to monitor energy consumption of CPU and GPU devices and estimate equivalent carbon emissions taking into account the regional carbon emissions accounting. The eco2AI library currently includes the largest and permanently enriched and maintained database of regional carbon emissions accounting based on the public available data in 209 countries. |
| Carbontracker | Carbontracker [106] is an open-source Python package for tracking and predicting the energy consumption and carbon footprint of computational workloads. The training of models can be cancelled as soon as the limit of environmental costs set by the user has been exceeded. |
| CodeCarbon | CodeCarbon [107] is an open-source software package that can easily be integrated in the Python codebase. It estimates the amount of $CO_2$ generated by the cloud or personal computing resources used to run the code. It further shows how the code can be optimized to reduce emissions and provides suggestions for hosting a cloud infrastructure in geographic regions that use renewable energy sources. |
| Energy Usage Reports | It is an open-source Python package [101] for calculating the energy and carbon emissions of algorithms. It provides an energy usage report in which the results are put into a context that is more understandable, such as car kilometers traveled. (This package is no longer actively maintained since it has been integrated into the tool CarbonCode). |
| EnergyVis | EnergyVis [108] is an interactive energy consumption tracker for ML models. It can be used for tracking, visualizing, and comparing model energy consumption in terms of energy consumption and $CO_2$ metrics. It further shows alternative deployment locations and hardware that can reduce the carbon footprint. |
| Experiment-Impact-Tracker | The experiment-impact-tracker [102] is an open-source drop-in method to track energy usage, carbon emissions, and compute utilization of the system used. It also generates carbon impact statements for enabling standardized reporting. |
| Green Algorithms | Green Algorithms [109] is a self-reporting free online tool that allows users to estimate the carbon footprint of their computations. The tool requires minimal information and considers a wide range of hardware configurations. After providing all the required information, the tool visualizes and contextualizes the estimated carbon footprint of the computations. |
| ML CO2 Impact | This free online tool relies on self-reporting [110]. The tool can estimate the carbon footprint of GPU computation by specifying hardware type, hours used, cloud provider, and region. |

One important issue that needs to be kept in mind when evaluating energy-efficient models is accuracy. If two models are compared and one is significantly more energy-efficient than the other, but at the same time has significantly lower accuracy, the comparison is not fair. Therefore, for example, Lee et al. [12] and Li et al. [104] include accuracy in their metrics. Furthermore, the optimal metric for energy efficiency should allow a fair comparison between different models. Therefore, this metric should ideally be stable under different influencing factors such as infrastructure. FLOPs could be an interesting metric, since it is nearly independent of hardware and software settings. Xu et al. [6] and Douwes et al. [20] suggest reporting the FLOPs during model training and inference. However, they would also encourage researchers to specify the total FLOPs during all experiments, including but not limited to parameter tuning and base implementation, to identify the wasted and redundant computations required to develop a new model/algorithm.

## 5. Discussion

This paper gives a holistic overview of techniques and approaches along the DL lifecycle to reduce the energy consumption of DL. Since it is an overview, the approaches are not described in detail. However, relevant references with more information are listed for each approach. It must be emphasized that the analyzed overviews are comprehensive, but not exhaustive. Due to the fact that the overviews focusing on resource-limited devices were not included, there are existing energy-efficient approaches (f. e. federated learning) that are not listed in this overview. However, this overview will hopefully give practitioners an idea at which point along the DL lifecycle and how the energy consumption of DL models can be reduced.

## 6. Conclusion

There are various approaches to increase the energy efficiency of DL. Such approaches have been extracted from recent overviews and classified within the six defined phases of the DL lifecycle. During the classification process, it was noticeable that approaches for the evaluation phase were not listed to the same extent in the overviews. This is because this phase is mainly about evaluating the energy efficiency of DL, not improving it. However, evaluating the energy efficiency of DL can also lead to improved energy efficiency, as practitioners' awareness about emitting $CO_2$ increases and some evaluation tools even provide recommendations on how to reduce energy consumption. The reporting of the models' energy metrics can also help achieve more energy efficiency in further research, as researchers can compare the models concerning their energy efficiency and thus, for example, use the model with lower energy consumption as a base or pre-trained model.

## 7. Future Work

Looking to the future of energy-efficient DL, it would be important to continue research in this direction, to develop more efficient hardware and software, more efficient architectures, and more or improved approaches reducing energy consumption. In addition, a standardized metric for energy efficiency should be introduced to enable the evaluation and comparison of DL models and to raise awareness among researchers and practitioners. Moreover, a clear and transparent reporting of the measurements is crucial. Here, not only the energy consumption of training but also that of the inference should be considered. Further research could include defining guidelines, proposing a process model or a framework that can serve as a step-by-step guide for practitioners to achieve energy improvements throughout the DL lifecycle.

## References

[1] T. Choudhary, V. Mishra, A. Goswami, and J. Sarangapani, 'A comprehensive survey on model compression and acceleration', *Artif. Intell. Rev.*, vol. 53, no. 7, pp. 5113–5155, Oct. 2020, doi: 10.1007/s10462-020-09816-7.

[2] S. Wang, 'Efficient deep learning', *Nat. Comput. Sci.*, vol. 1, no. 3, pp. 181–182, Mar. 2021, doi: 10.1038/s43588-021-00042-x.

[3] E. Strubell, A. Ganesh, and A. McCallum, 'Energy and Policy Considerations for Deep Learning in NLP', Jan. 2019, doi: 10.48550/arXiv.1906.02243.

[4] M. Kumar, X. Zhang, L. Liu, Y. Wang, and W. Shi, 'Energy-Efficient Machine Learning on the Edges', in *2020 IEEE International Parallel and Distributed Processing Symposium Workshops (IPDPSW)*, New Orleans, LA, USA, May 2020, pp. 912–921. doi: 10.1109/IPDPSW50202.2020.00153.

[5] G. Menghani, 'Efficient Deep Learning: A Survey on Making Deep Learning Models Smaller, Faster, and Better', Jun. 2021. [Online]. Available: http://arxiv.org/pdf/2106.08962v2

[6] C.-J. Wu *et al.*, 'Sustainable AI: Environmental Implications, Challenges and Opportunities', 2021, doi: 10.48550/ARXIV.2111.00364.

[7] M. Schulz *et al.*, 'DASC-PM v1.0 - Ein Vorgehensmodell für Data-Science-Projekte', Feb. 2021, doi: 10.25673/32872.2.

[8] R. Wirth and J. Hipp, 'CRISP-DM: Towards a Standard Process Model for Data Mining.' 2000.

[9] J. Xu, W. Zhou, Z. Fu, H. Zhou, and L. Li, 'A Survey on Green Deep Learning', Nov. 2021. [Online]. Available: http://arxiv.org/pdf/2111.05193v2

[10] H. Cai *et al.*, 'Enable Deep Learning on Mobile Devices: Methods, Systems, and Applications', *ACM Trans. Des. Autom. Electron. Syst.*, vol. 27, no. 3, pp. 1–50, May 2022, doi: 10.1145/3486618.

[11] D. Liu, H. Kong, X. Luo, W. Liu, and R. Subramaniam, 'Bringing AI to edge: From deep learning's perspective', *Neurocomputing*, vol. 485, pp. 297–320, May 2022, doi: 10.1016/j.neucom.2021.04.141.

[12] J. Lee *et al.*, 'Resource-Efficient Deep Learning: A Survey on Model-, Arithmetic-, and Implementation-Level Techniques'. arXiv, Dec. 30, 2021. Accessed: Jun. 14, 2022. [Online]. Available: http://arxiv.org/abs/2112.15131

[13] A. Vaswani *et al.*, 'Attention Is All You Need'. arXiv, Dec. 05, 2017. Accessed: Jul. 04, 2022. [Online]. Available: http://arxiv.org/abs/1706.03762

[14] H. Cai, C. Gan, T. Wang, Z. Zhang, and S. Han, 'Once-for-All: Train One Network and Specialize it for Efficient Deployment', *ArXiv190809791 Cs Stat*, Apr. 2020, Accessed: May 10, 2022. [Online]. Available: http://arxiv.org/abs/1908.09791

[15] A. Fan *et al.*, 'Training with Quantization Noise for Extreme Model Compression', *ArXiv Learn.*, Apr. 2020.

[16] N. Kitaev, Ł. Kaiser, and A. Levskaya, 'Reformer: The Efficient Transformer'. arXiv, Feb. 18, 2020. Accessed: Jun. 22, 2022. [Online]. Available: http://arxiv.org/abs/2001.04451

[17] Jingjing Xu, Liang Zhao, Junyang Lin, Rundong Gao, Xu Sun, and Hongxia Yang, 'KNAS: Green Neural Architecture Search', *ICML*, 2021.

[18] S. Cahyawijaya *et al.*, 'Greenformer: Factorization Toolkit for Efficient Deep Neural Networks', *ArXiv Learn.*, Sep. 2021.

[19] T. Hoefler, D. Alistarh, T. Ben-Nun, N. Dryden, and A. Peste, 'Sparsity in Deep Learning: Pruning and growth for efficient inference and training in neural networks', *ArXiv Learn.*, 2021.

[20] C. Douwes, P. Esling, and J.-P. Briot, 'Energy Consumption of Deep Generative Audio Models'. arXiv, Oct. 13, 2021. Accessed: Jul. 06, 2022. [Online]. Available: http://arxiv.org/abs/2107.02621

[21] David A. Patterson *et al.*, 'Carbon Emissions and Large Neural Network Training', *ArXiv*, 2021.

[22] R. Schwartz, J. Dodge, N. A. Smith, and O. Etzioni, 'Green AI', Jul. 2019. [Online]. Available: http://arxiv.org/pdf/1907.10597v3

[23] A. van Wynsberghe, 'Sustainable AI: AI for sustainability and the sustainability of AI', *AI Ethics*, vol. 1, no. 3, pp. 213–218, Jan. 2021, doi: 10.1007/s43681-021-00043-6.

[24] S. R. Young *et al.*, 'Evolving Energy Efficient Convolutional Neural Networks', in *2019 IEEE International Conference on Big Data (Big Data)*, Los Angeles, CA, USA, Dec. 2019, pp. 4479–4485. doi: 10.1109/BigData47090.2019.9006239.

[25] Martín Abadi, Paul Barham, Jianmin Chen, Zhifeng Chen, Andy Davis, Jeffrey Dean, Matthieu Devin, Sanjay Ghemawat, Geoffrey Irving, Michael Isard, Manjunath Kudlur, Josh Levenberg, Rajat Monga, Sherry Moore, Derek G. Murray, Benoit Steiner, Paul Tucker, and Vijay Vasudevan, Pete Warden, Martin Wicke, Yuan Yu, and Xiaoqiang Zheng, 'TensorFlow: A System for Large-Scale Machine Learning', *12th USENIX Symp. Oper. Syst. Des. Implement. OSDI 16*, pp. 265–283, 2016.

[26] A. Paszke *et al.*, 'PyTorch: An Imperative Style, High-Performance Deep Learning Library'. arXiv, Dec. 03, 2019. Accessed: Jun. 25, 2022. [Online]. Available: http://arxiv.org/abs/1912.01703

[27] R. Raina, A. Madhavan, and A. Y. Ng, 'Large-scale deep unsupervised learning using graphics processors', in *Proceedings of the 26th Annual International Conference on Machine Learning - ICML '09*, Montreal, Quebec, Canada, 2009, pp. 1–8. doi: 10.1145/1553374.1553486.

[28] N. P. Jouppi *et al.*, 'In-Datacenter Performance Analysis of a Tensor Processing Unit', in *Proceedings of the 44th Annual International Symposium on Computer Architecture*, Toronto ON Canada, Jun. 2017, pp. 1–12. doi: 10.1145/3079856.3080246.

[29] C. D. Schuman *et al.*, 'A Survey of Neuromorphic Computing and Neural Networks in Hardware'. arXiv, May 19, 2017. Accessed: Jun. 26, 2022. [Online]. Available: http://arxiv.org/abs/1705.06963

[30] A. Shrestha, H. Fang, Z. Mei, D. P. Rider, Q. Wu, and Q. Qiu, 'A Survey on Neuromorphic Computing: Models and Hardware', *IEEE Circuits Syst. Mag.*, vol. 22, no. 2, pp. 6–35, 2022, doi: 10.1109/MCAS.2022.3166331.

[31] E. D. Cubuk, B. Zoph, J. Shlens, and Q. V. Le, 'Randaugment: Practical automated data augmentation with a reduced search space', in *2020 IEEE/CVF Conference on Computer Vision and Pattern Recognition Workshops (CVPRW)*, Seattle, WA, USA, Jun. 2020, pp. 3008–3017. doi: 10.1109/CVPRW50498.2020.00359.

[32] X. Wang, H. Pham, Z. Dai, and G. Neubig, 'SwitchOut: an Efficient Data Augmentation Algorithm for Neural Machine Translation'. arXiv, Aug. 28, 2018. Accessed: Jul. 06, 2022. [Online]. Available: http://arxiv.org/abs/1808.07512

[33] S. Mohamadi and H. Amindavar, 'Deep Bayesian Active Learning, A Brief Survey on Recent Advances'. arXiv, Apr. 21, 2022. Accessed: Jun. 13, 2022. [Online]. Available: http://arxiv.org/abs/2012.08044

[34] P. Ren *et al.*, 'A Survey of Deep Active Learning', *ACM Comput. Surv.*, vol. 54, no. 9, pp. 1–40, Dec. 2022, doi: 10.1145/3472291.

[35] P. F. Jacobs, G. M. de B. Wenniger, M. Wiering, and L. Schomaker, 'Active learning for reducing labeling effort in text classification tasks'. Nov. 03, 2021. Accessed: Jun. 15, 2022. [Online]. Available: http://arxiv.org/abs/2109.04847

[36] K. Margatina, L. Barrault, and N. Aletras, 'Bayesian Active Learning with Pretrained Language Models', Jan. 2021, Accessed: Jun. 15, 2022. [Online]. Available: https://openreview.net/forum?id=oO1QGIKkEsY

[37] X. Qiu, T. Sun, Y. Xu, Y. Shao, N. Dai, and X. Huang, 'Pre-trained models for natural language processing: A survey', *Sci. China Technol. Sci.*, vol. 63, no. 10, pp. 1872–1897, Oct. 2020, doi: 10.1007/s11431-020-1647-3.

[38] F. Chollet, 'Xception: Deep Learning with Depthwise Separable Convolutions'. arXiv, Apr. 04, 2017. Accessed: Jun. 22, 2022. [Online]. Available: http://arxiv.org/abs/1610.02357

[39] Z. Qin, Z. Zhang, X. Chen, C. Wang, and Y. Peng, 'Fd-Mobilenet: Improved Mobilenet with a Fast Downsampling Strategy', in *2018 25th IEEE International Conference on Image Processing (ICIP)*, Athens, Oct. 2018, pp. 1363–1367. doi: 10.1109/ICIP.2018.8451355.

[40] J. Jin, A. Dundar, and E. Culurciello, 'Flattened Convolutional Neural Networks for Feedforward Acceleration'. arXiv, Nov. 20, 2015. Accessed: Jun. 22, 2022. [Online]. Available: http://arxiv.org/abs/1412.5474

[41] G. Huang, Z. Liu, L. van der Maaten, and K. Q. Weinberger, 'Densely Connected Convolutional Networks'. arXiv, Jan. 28, 2018. Accessed: Jun. 22, 2022. [Online]. Available: http://arxiv.org/abs/1608.06993

[42] G. Huang, S. Liu, L. van der Maaten, and K. Q. Weinberger, 'CondenseNet: An Efficient DenseNet using Learned Group Convolutions'. arXiv, Jun. 07, 2018. Accessed: Jun. 22, 2022. [Online]. Available: http://arxiv.org/abs/1711.09224

[43] M. Sandler, A. Howard, M. Zhu, A. Zhmoginov, and L.-C. Chen, 'MobileNetV2: Inverted Residuals and Linear Bottlenecks'. arXiv, Mar. 21, 2019. Accessed: Jun. 22, 2022. [Online]. Available: http://arxiv.org/abs/1801.04381

[44] A. G. Howard *et al.*, 'MobileNets: Efficient Convolutional Neural Networks for Mobile Vision Applications'. arXiv, Apr. 16, 2017. Accessed: Jun. 22, 2022. [Online]. Available: http://arxiv.org/abs/1704.04861

[45] F. N. Iandola, S. Han, M. W. Moskewicz, K. Ashraf, W. J. Dally, and K. Keutzer, 'SqueezeNet: AlexNet-level accuracy with 50x fewer parameters and <0.5MB model size'. arXiv, Nov. 04, 2016. Accessed: Jun. 22, 2022. [Online]. Available: http://arxiv.org/abs/1602.07360

[46] Z. Dai, Z. Yang, Y. Yang, J. Carbonell, Q. V. Le, and R. Salakhutdinov, 'Transformer-XL: Attentive Language Models Beyond a Fixed-Length Context'. arXiv, Jun. 02, 2019. Accessed: Jun. 22, 2022. [Online]. Available: http://arxiv.org/abs/1901.02860

[47] G. M. Correia, V. Niculae, and A. F. T. Martins, 'Adaptively Sparse Transformers'. arXiv, Sep. 06, 2019. Accessed: Jun. 22, 2022. [Online]. Available: http://arxiv.org/abs/1909.00015

[48] R. Child, S. Gray, A. Radford, and I. Sutskever, 'Generating Long Sequences with Sparse Transformers'. arXiv, Apr. 23, 2019. Accessed: Jun. 22, 2022. [Online]. Available: http://arxiv.org/abs/1904.10509

[49] K. Choromanski *et al.*, 'Rethinking Attention with Performers'. arXiv, Mar. 09, 2021. Accessed: Jun. 22, 2022. [Online]. Available: http://arxiv.org/abs/2009.14794

[50] M. R. Costa-Jussà and J. A. R. Fonollosa, 'Character-based Neural Machine Translation'. arXiv, Jun. 30, 2016. Accessed: Jun. 24, 2022. [Online]. Available: http://arxiv.org/abs/1603.00810

[51] R. Sennrich, B. Haddow, and A. Birch, 'Neural Machine Translation of Rare Words with Subword Units'. arXiv, Jun. 10, 2016. Accessed: Jun. 24, 2022. [Online]. Available: http://arxiv.org/abs/1508.07909

[52] Q. Liu *et al.*, 'Hierarchical Softmax for End-to-End Low-resource Multilingual Speech Recognition'. arXiv, Apr. 08, 2022. Accessed: Jul. 12, 2022. [Online]. Available: http://arxiv.org/abs/2204.03855

[53] W. Chen, D. Grangier, and M. Auli, 'Strategies for Training Large Vocabulary Neural Language Models'. arXiv, Dec. 15, 2015. Accessed: Jun. 24, 2022. [Online]. Available: http://arxiv.org/abs/1512.04906

[54] S. Jean, K. Cho, R. Memisevic, and Y. Bengio, 'On Using Very Large Target Vocabulary for Neural Machine Translation'. arXiv, Mar. 18, 2015. Accessed: Jun. 24, 2022. [Online]. Available: http://arxiv.org/abs/1412.2007

[55] S. Vijayanarasimhan, J. Shlens, R. Monga, and J. Yagnik, 'Deep Networks With Large Output Spaces'. arXiv, Apr. 10, 2015. Accessed: Jun. 24, 2022. [Online]. Available: http://arxiv.org/abs/1412.7479

[56] J. Devlin, R. Zbib, Z. Huang, T. Lamar, R. Schwartz, and J. Makhoul, 'Fast and Robust Neural Network Joint Models for Statistical Machine Translation', in *Proceedings of the 52nd Annual Meeting of the Association for Computational Linguistics (Volume 1: Long Papers)*, Baltimore, Maryland, 2014, pp. 1370–1380. doi: 10.3115/v1/P14-1129.

[57] R. Shu and H. Nakayama, 'Compressing Word Embeddings via Deep Compositional Code Learning'. arXiv, Nov. 17, 2017. Accessed: Jun. 24, 2022. [Online]. Available: http://arxiv.org/abs/1711.01068

[58] M. Joshi, E. Choi, O. Levy, D. S. Weld, and L. Zettlemoyer, 'pair2vec: Compositional Word-Pair Embeddings for Cross-Sentence Inference'. arXiv, Apr. 05, 2019. Accessed: Jun. 24, 2022. [Online]. Available: http://arxiv.org/abs/1810.08854

[59] H.-J. M. Shi, D. Mudigere, M. Naumov, and J. Yang, 'Compositional Embeddings Using Complementary Partitions for Memory-Efficient Recommendation Systems', in *Proceedings of the 26th ACM SIGKDD International Conference on Knowledge Discovery & Data Mining*, Virtual Event CA USA, Aug. 2020, pp. 165–175. doi: 10.1145/3394486.3403059.

[60] L. Mou, R. Jia, Y. Xu, G. Li, L. Zhang, and Z. Jin, 'Distilling Word Embeddings: An Encoding Approach', in *Proceedings of the 25th ACM International on Conference on Information and Knowledge Management*, Indianapolis Indiana USA, Oct. 2016, pp. 1977–1980. doi: 10.1145/2983323.2983888.

[61] P. H. Chen, S. Si, Y. Li, C. Chelba, and C. Hsieh, 'GroupReduce: Block-Wise Low-Rank Approximation for Neural Language Model Shrinking'. arXiv, Jun. 18, 2018. Accessed: Jul. 12, 2022. [Online]. Available: http://arxiv.org/abs/1806.06950

[62] C. Wang, K. Cho, and J. Gu, 'Neural Machine Translation with Byte-Level Subwords', *Proc. AAAI Conf. Artif. Intell.*, vol. 34, no. 05, pp. 9154–9160, Apr. 2020, doi: 10.1609/aaai.v34i05.6451.

[63] G. Pleiss, D. Chen, G. Huang, T. Li, L. van der Maaten, and K. Q. Weinberger, 'Memory-Efficient Implementation of DenseNets'. arXiv, Jul. 21, 2017. Accessed: Jun. 24, 2022. [Online]. Available: http://arxiv.org/abs/1707.06990

[64] T. Chen, B. Xu, C. Zhang, and C. Guestrin, 'Training Deep Nets with Sublinear Memory Cost'. arXiv, Apr. 22, 2016. Accessed: Jun. 24, 2022. [Online]. Available: http://arxiv.org/abs/1604.06174

[65] L. Wang *et al.*, 'Superneurons: dynamic GPU memory management for training deep neural networks', in *Proceedings of the 23rd ACM SIGPLAN Symposium on Principles and Practice of Parallel Programming*, Vienna Austria, Feb. 2018, pp. 41–53. doi: 10.1145/3178487.3178491.

[66] S. Rajbhandari, J. Rasley, O. Ruwase, and Y. He, 'ZeRO: Memory optimizations Toward Training Trillion Parameter Models', in *SC20: International Conference for High Performance Computing, Networking, Storage and Analysis*, Atlanta, GA, USA, Nov. 2020, pp. 1–16. doi: 10.1109/SC41405.2020.00024.

[67] M. Dehghani, S. Gouws, O. Vinyals, J. Uszkoreit, and Ł. Kaiser, 'Universal Transformers'. arXiv, Mar. 05, 2019. Accessed: Jun. 24, 2022. [Online]. Available: http://arxiv.org/abs/1807.03819

[68] Z. Lan, M. Chen, S. Goodman, K. Gimpel, P. Sharma, and R. Soricut, 'ALBERT: A Lite BERT for Self-supervised Learning of Language Representations'. arXiv, Feb. 08, 2020. Accessed: Jun. 24, 2022. [Online]. Available: http://arxiv.org/abs/1909.11942

[69] S. Takase and S. Kiyono, 'Lessons on Parameter Sharing across Layers in Transformers'. arXiv, Apr. 20, 2022. Accessed: Jun. 24, 2022. [Online]. Available: http://arxiv.org/abs/2104.06022

[70] K. Hashimoto, C. Xiong, Y. Tsuruoka, and R. Socher, 'A Joint Many-Task Model: Growing a Neural Network for Multiple NLP Tasks'. arXiv, Jul. 24, 2017. Accessed: Jun. 24, 2022. [Online]. Available: http://arxiv.org/abs/1611.01587

[71] E. Park *et al.*, 'Big/little deep neural network for ultra low power inference', in *2015 International Conference on Hardware/Software Codesign and System Synthesis (CODES+ISSS)*, Amsterdam, Netherlands, Oct. 2015, pp. 124–132. doi: 10.1109/CODESISSS.2015.7331375.

[72] A. Gormez and E. Koyuncu, 'Class Means as an Early Exit Decision Mechanism'. arXiv, Oct. 20, 2021. Accessed: Jun. 25, 2022. [Online]. Available: http://arxiv.org/abs/2103.01148

[73] L. Yang, Y. Han, X. Chen, S. Song, J. Dai, and G. Huang, 'Resolution Adaptive Networks for Efficient Inference'. arXiv, May 18, 2020. Accessed: Jun. 25, 2022. [Online]. Available: http://arxiv.org/abs/2003.07326

[74] X. Wang, F. Yu, Z.-Y. Dou, T. Darrell, and J. E. Gonzalez, 'SkipNet: Learning Dynamic Routing in Convolutional Networks'. arXiv, Jul. 25, 2018. Accessed: Jun. 25, 2022. [Online]. Available: http://arxiv.org/abs/1711.09485

[75] Z. Wu *et al.*, 'BlockDrop: Dynamic Inference Paths in Residual Networks'. arXiv, Jan. 28, 2019. Accessed: Jun. 25, 2022. [Online]. Available: http://arxiv.org/abs/1711.08393

[76] A. Veit and S. Belongie, 'Convolutional Networks with Adaptive Inference Graphs'. arXiv, May 08, 2020. Accessed: Jun. 25, 2022. [Online]. Available: http://arxiv.org/abs/1711.11503

[77] D. Lepikhin *et al.*, 'GShard: Scaling Giant Models with Conditional Computation and Automatic Sharding'. arXiv, Jun. 30, 2020. Accessed: Jun. 25, 2022. [Online]. Available: http://arxiv.org/abs/2006.16668

[78] J. Lin *et al.*, 'M6: A Chinese Multimodal Pretrainer'. arXiv, May 29, 2021. Accessed: Jun. 25, 2022. [Online]. Available: http://arxiv.org/abs/2103.00823

[79] J.-X. Mi, J. Feng, and K.-Y. Huang, 'Designing efficient convolutional neural network structure: A survey', *Neurocomputing*, vol. 489, pp. 139–156, Jun. 2022, doi: 10.1016/j.neucom.2021.08.158.

[80] E. Real, A. Aggarwal, Y. Huang, and Q. V. Le, 'Regularized Evolution for Image Classifier Architecture Search', *Proc. AAAI Conf. Artif. Intell.*, vol. 33, pp. 4780–4789, Jul. 2019, doi: 10.1609/aaai.v33i01.33014780.

[81] Y. Jiang, C. Hu, T. Xiao, C. Zhang, and J. Zhu, 'Improved Differentiable Architecture Search for Language Modeling and Named Entity Recognition', in *Proceedings of the 2019 Conference on Empirical Methods in Natural Language Processing and the 9th International Joint*

[81] *Conference on Natural Language Processing (EMNLP-IJCNLP)*, Hong Kong, China, 2019, pp. 3583–3588. doi: 10.18653/v1/D19-1367.

[82] A. Zela, T. Elsken, T. Saikia, Y. Marrakchi, T. Brox, and F. Hutter, 'Understanding and Robustifying Differentiable Architecture Search'. arXiv, Jan. 28, 2020. Accessed: Jun. 25, 2022. [Online]. Available: http://arxiv.org/abs/1909.09656

[83] X. Chu and B. Zhang, 'Noisy Differentiable Architecture Search'. arXiv, Oct. 17, 2021. Accessed: Jun. 25, 2022. [Online]. Available: http://arxiv.org/abs/2005.03566

[84] R. Luo, F. Tian, T. Qin, E. Chen, and T.-Y. Liu, 'Neural Architecture Optimization'. arXiv, Sep. 04, 2019. Accessed: Jul. 12, 2022. [Online]. Available: http://arxiv.org/abs/1808.07233

[85] W. Chen, X. Gong, and Z. Wang, 'Neural Architecture Search on ImageNet in Four GPU Hours: A Theoretically Inspired Perspective'. arXiv, Mar. 15, 2021. Accessed: Jun. 25, 2022. [Online]. Available: http://arxiv.org/abs/2102.11535

[86] M. S. Abdelfattah, A. Mehrotra, Ł. Dudziak, and N. D. Lane, 'Zero-Cost Proxies for Lightweight NAS'. arXiv, Mar. 19, 2021. Accessed: Jun. 25, 2022. [Online]. Available: http://arxiv.org/abs/2101.08134

[87] Z. Alyafeai, M. S. AlShaibani, and I. Ahmad, 'A Survey on Transfer Learning in Natural Language Processing'. arXiv, May 31, 2020. Accessed: Jun. 15, 2022. [Online]. Available: http://arxiv.org/abs/2007.04239

[88] F. Zhuang *et al.*, 'A Comprehensive Survey on Transfer Learning', *Proc. IEEE*, vol. 109, no. 1, pp. 43–76, Jan. 2021, doi: 10.1109/JPROC.2020.3004555.

[89] S. Ioffe and C. Szegedy, 'Batch Normalization: Accelerating Deep Network Training by Reducing Internal Covariate Shift'. arXiv, Mar. 02, 2015. Accessed: Jul. 12, 2022. [Online]. Available: http://arxiv.org/abs/1502.03167

[90] J. L. Ba, J. R. Kiros, and G. E. Hinton, 'Layer Normalization'. arXiv, Jul. 21, 2016. Accessed: Jul. 03, 2022. [Online]. Available: http://arxiv.org/abs/1607.06450

[91] Y. Wu and K. He, 'Group Normalization'. arXiv, Jun. 11, 2018. Accessed: Jul. 03, 2022. [Online]. Available: http://arxiv.org/abs/1803.08494

[92] L. Gong, D. He, Z. Li, T. Qin, L. Wang, and T.-Y. Liu, 'Efficient Training of BERT by Progressively Stacking', presented at the Proceedings of the 36 th International Conference on Machine Learnin, Long Beach, California, 2019, vol. PMLR 97.

[93] C. Yang, S. Wang, C. Yang, Y. Li, R. He, and J. Zhang, 'Progressively Stacking 2.0: A Multi-stage Layerwise Training Method for BERT Training Speedup'. arXiv, Nov. 27, 2020. Accessed: Jul. 03, 2022. [Online]. Available: http://arxiv.org/abs/2011.13635

[94] X. He, F. Xue, X. Ren, and Y. You, 'Large-Scale Deep Learning Optimizations: A Comprehensive Survey'. arXiv, Nov. 01, 2021. Accessed: Jun. 09, 2022. [Online]. Available: http://arxiv.org/abs/2111.00856

[95] P. Micikevicius *et al.*, 'Mixed Precision Training'. arXiv, Feb. 15, 2018. Accessed: Jun. 30, 2022. [Online]. Available: http://arxiv.org/abs/1710.03740

[96] D. Ghimire, D. Kil, and S. Kim, 'A Survey on Efficient Convolutional Neural Networks and Hardware Acceleration', *Electronics*, vol. 11, no. 6, p. 945, Mar. 2022, doi: 10.3390/electronics11060945.

[97] A. Gholami, S. Kim, Z. Dong, Z. Yao, M. W. Mahoney, and K. Keutzer, 'A Survey of Quantization Methods for Efficient Neural Network Inference', *ArXiv Comput. Vis. Pattern Recognit.*, 2021.

[98] J. Gou, B. Yu, S. J. Maybank, and D. Tao, 'Knowledge Distillation: A Survey', *Int. J. Comput. Vis.*, vol. 129, no. 6, pp. 1789–1819, Jun. 2021, doi: 10.1007/s11263-021-01453-z.

[99] J. Yu, L. Yang, N. Xu, J. Yang, and T. S. Huang, 'Slimmable Neural Networks.', *ArXiv Comput. Vis. Pattern Recognit.*, Dec. 2018.

[100] A. Fan, E. Grave, and A. Joulin, 'Reducing Transformer Depth on Demand with Structured Dropout'. arXiv, Sep. 25, 2019. Accessed: Jun. 24, 2022. [Online]. Available: http://arxiv.org/abs/1909.11556

[101] K. Lottick, S. Susai, S. A. Friedler, and J. P. Wilson, 'Energy Usage Reports: Environmental awareness as part of algorithmic accountability', *ArXiv Learn.*, Nov. 2019.

[102] P. Henderson, J. Hu, J. Romoff, E. Brunskill, D. Jurafsky, and J. Pineau, 'Towards the Systematic Reporting of the Energy and Carbon Footprints of Machine Learning', Jan. 2020. [Online]. Available: http://arxiv.org/pdf/2002.05651v1

[103] W. A. Hanafy, T. Molom-Ochir, and R. Shenoy, 'Design Considerations for Energy-efficient Inference on Edge Devices', *E-Energy*, pp. 302–308, Jun. 2021, doi: 10.1145/3447555.3465326.

[104] B. Li *et al.*, 'Full-Cycle Energy Consumption Benchmark for Low-Carbon Computer Vision', *ArXiv210813465 Cs*, Oct. 2021, Accessed: May 10, 2022. [Online]. Available: http://arxiv.org/abs/2108.13465

[105] S. Budennyy *et al.*, 'Eco2AI: carbon emissions tracking of machine learning models as the first step towards sustainable AI'. arXiv, Aug. 03, 2022. Accessed: Sep. 26, 2022. [Online]. Available: http://arxiv.org/abs/2208.00406

[106] L. F. W. Anthony, B. Kanding, and R. Selvan, 'Carbontracker: Tracking and Predicting the Carbon Footprint of Training Deep Learning Models', *ArXiv200703051 Cs Eess Stat*, Jul. 2020, Accessed: May 10, 2022. [Online]. Available: http://arxiv.org/abs/2007.03051

[107] V. Schmidt *et al.*, *mlco2/codecarbon: v2.1.4*. Zenodo, 2022. doi: 10.5281/ZENODO.4658424.

[108] O. Shaikh *et al.*, 'EnergyVis: Interactively Tracking and Exploring Energy Consumption for ML Models', *ArXiv Learn.*, 2021.

[109] L. Lannelongue, J. Grealey, and M. Inouye, 'Green Algorithms: Quantifying the carbon emissions of computation.', Jul. 2020.

[110] A. Lacoste, Alexandra Luccioni, A. Luccioni, V. Schmidt, and T. Dandres, 'Quantifying the Carbon Emissions of Machine Learning.', *ArXiv Comput. Soc.*, Oct. 2019.